\documentclass{article} 
\usepackage{iclr2026/iclr2026_conference,times}

\usepackage{amsmath,amsfonts,bm}

\def\eqref#1{equation~\ref{#1}}

\def\1{\bm{1}}

\DeclareMathAlphabet{\mathsfit}{\encodingdefault}{\sfdefault}{m}{sl}
\SetMathAlphabet{\mathsfit}{bold}{\encodingdefault}{\sfdefault}{bx}{n}

\usepackage{hyperref}
\usepackage{url}

\usepackage{threeparttable}
\usepackage{booktabs}
\usepackage{multirow}
\usepackage{array}
\usepackage{graphicx}
\usepackage{adjustbox}
\usepackage{xcolor}
\usepackage{amssymb}
\usepackage{pifont}
\usepackage{subcaption}

\definecolor{myblue}{RGB}{0,64,128}    
\definecolor{myorange}{RGB}{200,100,0} 
\newcommand{\first}[1]{\textbf{\textcolor{myorange}{#1}}}
\newcommand{\second}[1]{\underline{\textcolor{myblue}{#1}}}
\definecolor{darkergreen}{RGB}{0,100,0}

\definecolor{rebuttalcolor}{RGB}{200,0,100}  
\usepackage{fontawesome5}

\title{PMDformer: Patch-Mean Decoupling Information Transformer for Long-term Forecasting}

\author{
\textbf{Ao Hu}$^{1,2}$\thanks{Equal contribution.} \quad 
\textbf{Liangjian Wen}$^{1,7}$\thanks{Corresponding author.} \quad 
\textbf{Jiang Duan}$^{1,6}$\footnotemark[1] \quad 
\textbf{Yong Dai}$^{5}$\thanks{Project leader.} \quad 
\textbf{HE YAN}$^{6}$ \quad
\textbf{Dongkai Wang}$^{1}$ \\
\textbf{Jun Wang}$^{1}$ \quad 
\textbf{Yukun Zhang}$^{4,2}$ \quad 
\textbf{Ruoxi Jiang}$^{2,3}$ \quad 
\textbf{Zenglin Xu}$^{2,3}$\footnotemark[2] \\[0.5em]
$^{1}$Southwestern University of Finance and Economics \quad
$^{2}$Shanghai Academy of AI for Science \\
$^{3}$Fudan University \quad
$^{4}$Harbin Institute of Technology, Shenzhen \quad
$^{5}$X-Humanoid Research Institute \\[0.3em]
$^{6}$Chengdu Everimaging Science and Technology Co., Ltd. \\
$^{7}$Artificial Intelligence and Digital Finance Key Laboratory of Sichuan Province \\
\texttt{\{huao1105, wlj6816, zenglin\}@gmail.com} \quad 
\texttt{duanj\_t@swufe.edu.cn}
}

\iclrfinalcopy 
\begin{document}

\thispagestyle{fancy}
\pagestyle{fancy}
\fancyhead{}
\lhead{Published as a conference paper at ICLR 2026}

\maketitle

\begin{abstract}
Long-term time series forecasting (LTSF) plays a crucial role in fields such as energy management, finance, and traffic prediction. Transformer-based models have adopted patch-based strategies to capture long-range dependencies, but accurately modeling shape similarities across patches and variables remains challenging due to scale differences. 
To address this, we introduce patch-mean decoupling (PMD), which separates the trend and residual shape information by subtracting the mean of each patch, preserving the original structure and ensuring that the {attention mechanism} captures true shape similarities. 
{Futhermore, to more effectively model long-range dependencies and capture cross-variable relationships}, we propose Trend Restoration Attention (TRA) and Proximal Variable Attention (PVA). The former module reintegrates the decoupled trend {from PMD while calculating attention output}. And the latter focuses {cross-variable} attention on the most relevant, recent time segments to avoid overfitting on outdated correlations. Combining these components, {we propose PMDformer, a model designed to effectively capture shape similarity in long-term forecasting scenarios. Extensive experiments indicate that PMDformer} outperforms existing state-of-the-art methods in stability and accuracy across multiple LTSF benchmarks. The code is available at \href{https://github.com/aohu1105/PMDformer}{\texttt{https://github.com/aohu1105/PMDformer}}.

\end{abstract}

\section{Introduction}
Long-term time series forecasting (LTSF) is a key task in machine learning, with wide applications in areas like energy management~\citep{energy_1}, financial markets~\citep{finance_1}, and traffic prediction~\citep{traffic_1, traffic_2}. Recent Transformer-based models have drawn inspiration from computer vision~\citep{VIT}, increasingly using patch-based strategies~\citep{PatchTST, Crossformer, Pathformer, timexer} to better capture long-range dependencies.
Most of these approaches treat variables independently (VI)~\citep{TimeBase, SparseTSF}, while variable-dependent (VD) methods~\citep{iTransformer, ModernTCN} that model interactions across variables have not yet shown clear gains over VI baselines.

\begin{figure*}[t] 
    \centering
    \begin{minipage}{0.63\textwidth} 
        \centering
        \includegraphics[width=\textwidth]{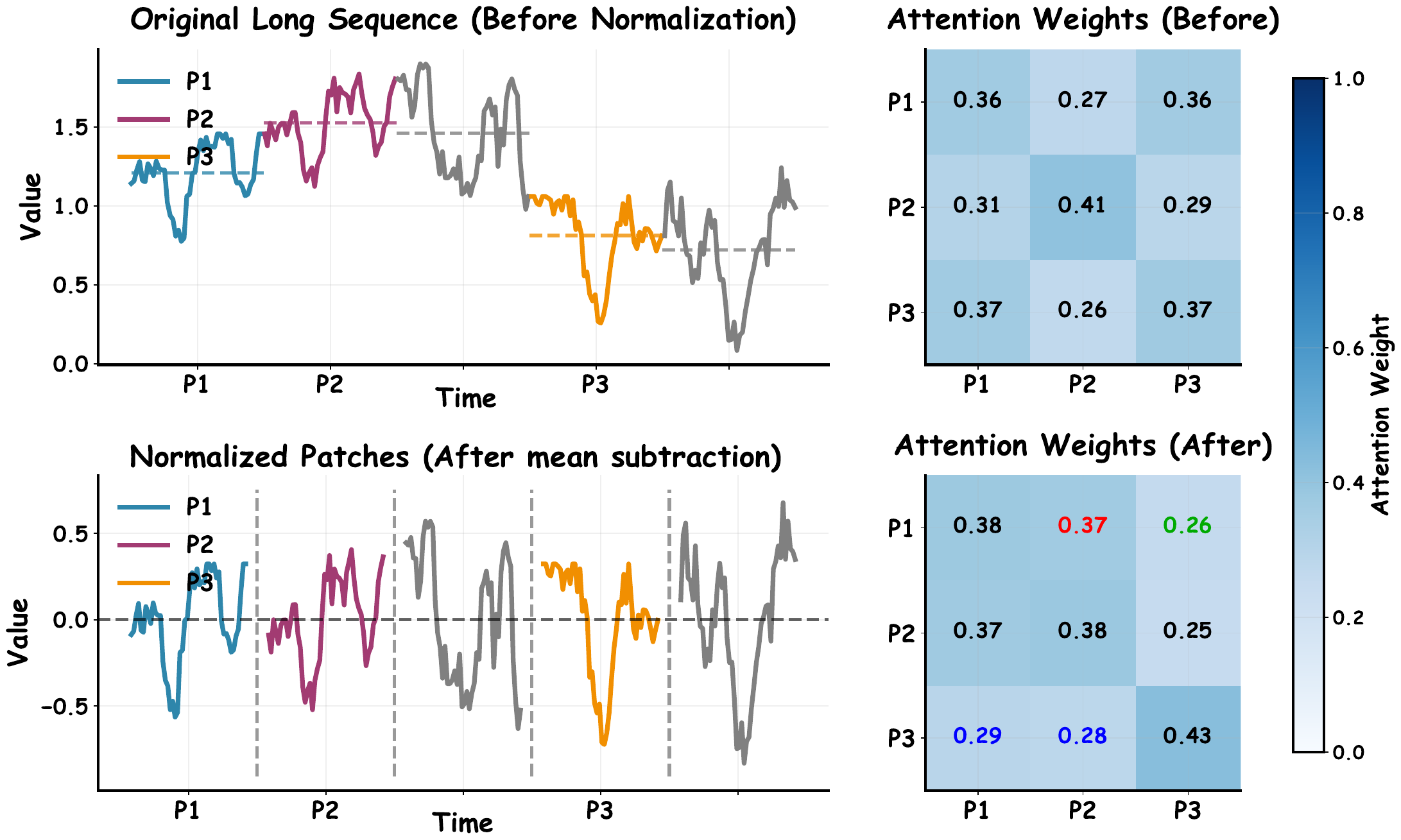}
        \caption{Attention weights for three patches before and after patch-mean decoupling. Scale differences initially obscure true shape similarity, {which are clearly revealed after decoupling as increased (\textcolor{red}{red}) or decreased (\textcolor{darkergreen}{green}) correlations, with analogous similarity shown in \textcolor{blue}{blue} for ($P_3$, $P_1$) and ($P_3$, $P_2$).}}\label{fig:attention_decoupling}  
    \end{minipage}
    \hfill 
    \begin{minipage}{0.35\textwidth} 
        \centering
        \vspace{0.4cm}
        \includegraphics[width=\textwidth]{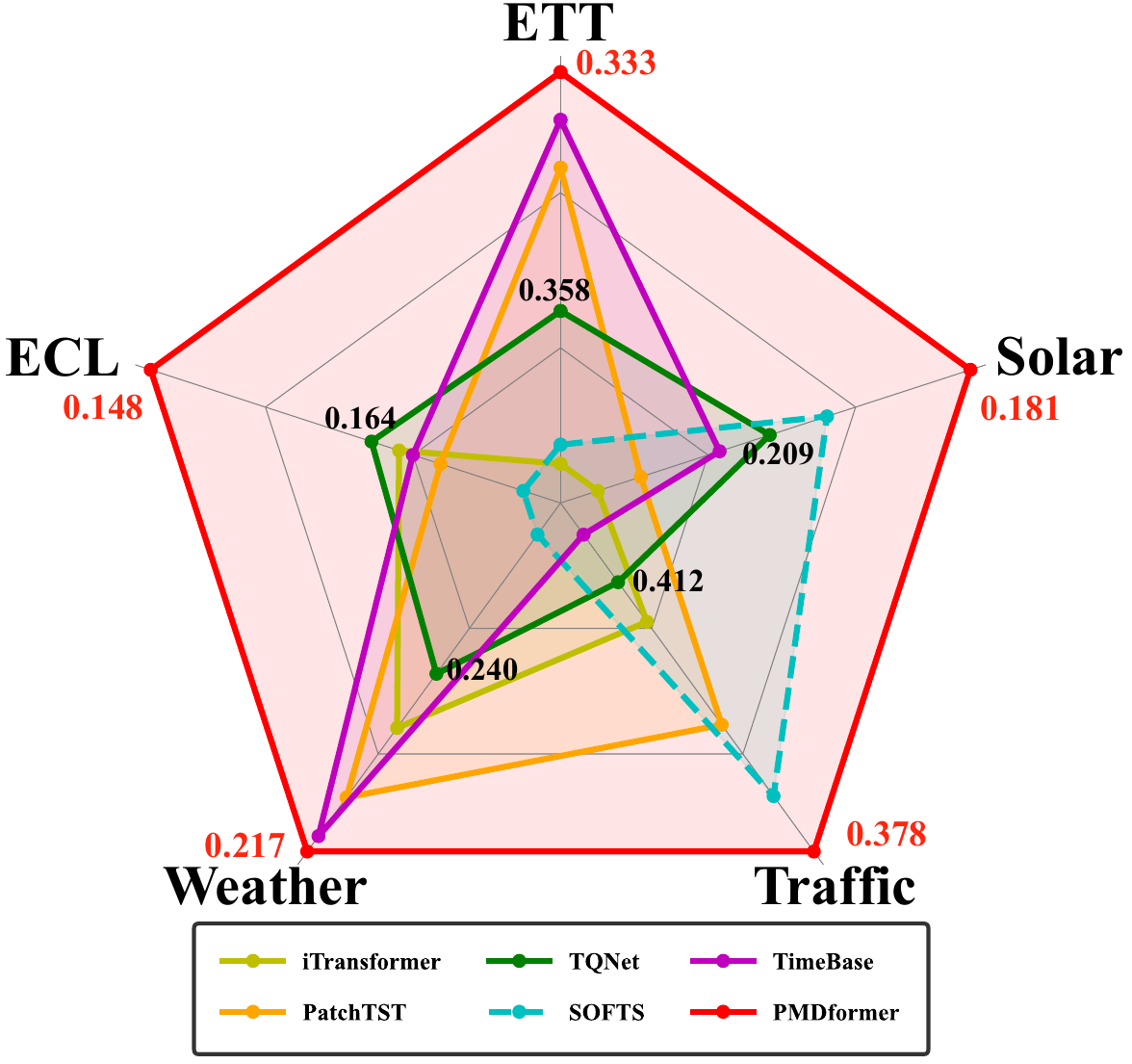} 
            \caption{Comparison of the MSE of all baselines with our proposed PMDformer. The results are the averages for all prediction lengths.} \label{fig:two-images} 
    \end{minipage}
    \vspace{-0.4cm}
\end{figure*}

Unlike 2D images with a fixed spatial structure, time series are one-dimensional curve~\cite{shape_3, shape_4}, with the primary focus being on capturing shape similarities between patches or variables~\citep{shape_1, shape_2} as well as modeling long-range trend~\citep{RLinear}. For instance, two patches may share similar trends, such as gradual increases with comparable rates of change. Identifying such shape correspondence helps the model extract temporally consistent patterns and improves forecast accuracy. However, time series data is inherently non-stationary~\citep{DishTS,nonstationary_1}, where patch scales fluctuate wildly across time. As illustrated in the top panels of Figure~\ref{fig:attention_decoupling}, {The attention weight of $(P_1, P_3)$ is higher than that of $(P_1, P_2)$, despite the more similar shape between $P_1$ and $P_2$. This occurs due to the different scales among $P_1$, $P_2$ and $P_3$, which influence the attention weights, thereby failing to reflect true shape similarity. Consequently, the model may learn incorrect similarity relationships, leading to performance degradation.} Furthermore, this scale bias is even more pronounced when modeling dependencies between variables, further hindering the effectiveness of VD models.

To balance the scale differences of patches, recent methods have employed Patch Normalization~\citep{SAN}, which Z-score normalizes each patch by subtracting the mean and dividing by the standard deviation. However, the removal of the standard deviation inadvertently distorts the original shape of the patch. As a result, it hampers the model's ability to identify shape similarities across patches or variables. In this paper, we propose a simple yet effective alternative method called patch-mean decoupling (PMD). We subtract the mean of each patch, which recenters each patch to zero mean and explicitly separates the long-range trend component which is encoded in the means of patches from the residual shape information. Unlike Patch Normalization, our method preserves the original amplitude variations and maintains the intrinsic shape structure, ensuring that the model better captures true shape similarities across patches. As shown in Figure~\ref{fig:attention_decoupling}, through our method attention favors shape-aligned pairs ($P_1, P_2$) over {shape-unaligned} ($P_1, P_3$).

PMD thus enables more shape-focused attention across patches and variables, revealing true similarities obscured by scales. For cross-variable shape modeling, {existing methods~\citep{ModernTCN, Crossformer} often compute interactions over the entire historical window. However, cross-variable relationships are often non-stationary and evolve over time, so recent interactions are the most predictive of future dynamics. For example, in financial markets asset correlations often spike sharply during crises. Relying on the entire historical dependencies introduces substantial noise and redundancy, degrading performance.} To address this, we introduce proximal variable attention (PVA), which confines self-attention to the most recent patch—the segment most proximal to the prediction horizon. By capturing shape similarities among variables in this temporally relevant window, PVA minimizes noise from historical drifts and risk overfitting.

Complementarily, recentering via PMD inherently attenuates the long-term trend signal, potentially overlooking global dependencies. To restore this without disrupting shape matching between temporal patches, we propose trend restoration attention (TRA), which explicitly injects the decoupled means (long-range trend information) into the value pathway of the attention mechanism. This seamless integration allows the model to jointly encode local shape patterns and global trend yielding more stable forecasts.

Building on above, we propose \textbf{PMDformer}, which combines patch-mean decoupling (PMD) module, Proximal variable attention (PVA), trend reinsertion attention (TRA) and a projection layer for final forecasting. The comparison of predictive accuracy of our PMDformer and other state of the art models refer to Figure~\ref{fig:two-images}. Our contributions are:
\begin{itemize}
    \item We introduce a novel mechanism to decouple trend and residual shape within the attention module via residual mean deduction, {enabling more effectively capture shape similarity among temporal patches and varibles.}
    \item We introduce proximal variable attention, which focuses on the most recent patch to capture the most relevant {shape similarities}, mitigating overfitting.
    \item We demonstrate the effectiveness of our approach through extensive experiments on a variety of LTSF benchmarks, showing that PMDformer provides more stable and accurate forecasts than current state-of-the-art methods.
\end{itemize}

\section{Related Work}
Deep learning models have demonstrated remarkable performance in long-term time series forecasting. These models can be broadly divided into Transformer-based models~\cite{transformer, autoformer, Pyraformer, FEDformer}, MLP-based models~\cite{DLinear, RLinear, timemixer, FDNet}, GNN-based models~\cite{CrossGNN, fouriergnn} and CNN-based models~\cite{micn, TSLANet, TimeCNN}. 
\paragraph{Transformer-based time series models.}
The success of Transformers~\citep{transformer} in NLP has inspired their adaptation for LTSF to capture long-range dependencies. Early models treat series as token sequences with efficient attention: Informer~\citep{Informer} uses ProbSparse for complexity reduction; Pyraformer~\citep{Pyraformer} employs pyramidal attention; Autoformer~\citep{autoformer} adds decomposition; and FEDformer~\citep{FEDformer} incorporates frequency blocks. Yet, their efficacy is challenged by simple linear models~\citep{DLinear}, underscoring needs for better temporal modeling.

\paragraph{Patch-based time series models.}
Inspired by vision transformers~\citep{VIT}, recent works segment time series into overlapping or non-overlapping patches to bolster local semantic capture. Transformer-based examples include PatchTST~\citep{PatchTST}, which uses variable-independent shared encoders for temporal patch semantics (SOTA in LTSF), and Pathformer~\citep{Pathformer} with multi-scale patches and adaptive path selection for intra/inter-dependencies. MLP variants like TSMixer~\citep{TSMixer} and PatchMixer~\citep{PatchMixer} model patch relations via MLPs, while foundation models such as Moirai~\citep{Moirai}, Timer~\citep{timer}, TimesFM~\citep{TimesFM}, and LLM-based~\citep{S2LLM, time-llm}leverage patches for pretraining and cross-modal alignment. Recent TimeBase~\citep{TimeBase} employs orthogonalized patches to reduce redundancy for SOTA efficiency, which further underscores patches' success in LTSF modeling.

\paragraph{Patch-Normalization.}
Due to the non-stationary nature of time series, some works~\citep{DishTS, RevIN} apply normalization to mitigate scale discrepancies and stabilize distributions. Among them, Patch-level normalization works include SAN~\citep{SAN}, a model-agnostic framework that adaptively normalizes slices by removing non-stationarity for flexible forecasting, and SIN~\citep{SIN}, which selectively learns normalization parameters to maximize local invariance and global variability, enabling interpretable long-term predictions. However, these normalization methods distort intrinsic patch shapes by scaling with standard deviation, hindering true shape similarity capture. In contrast, our PMD overcomes through mean subtraction to preserve amplitudes.


\section{Proposed Method}
We consider the task of long-term time series forecasting, where the goal is to predict the future evolution of multiple correlated variables given their historical observations. Formally, let $\mathbf{X} = \{x_{t} \in \mathbb{R}^{C} \mid t = 1, 2, \ldots, L\}$
denote an input sequence of length $L$, where $C$ is the number of variables. Each $x_t = (x_t^1, x_t^2, \ldots, x_t^C)$ contains the values of all variables at time $t$. Given $\mathbf{X}$, the objective is to forecast the subsequent $T$ time steps $\hat{\mathbf{Y}} = \{\hat{x}_{t} \in \mathbb{R}^{C} \mid t = L+1, \ldots, L+T\}$.

\subsection{The General Structure}
\begin{figure*}[t]    
    \centering
    \includegraphics[width=\textwidth]{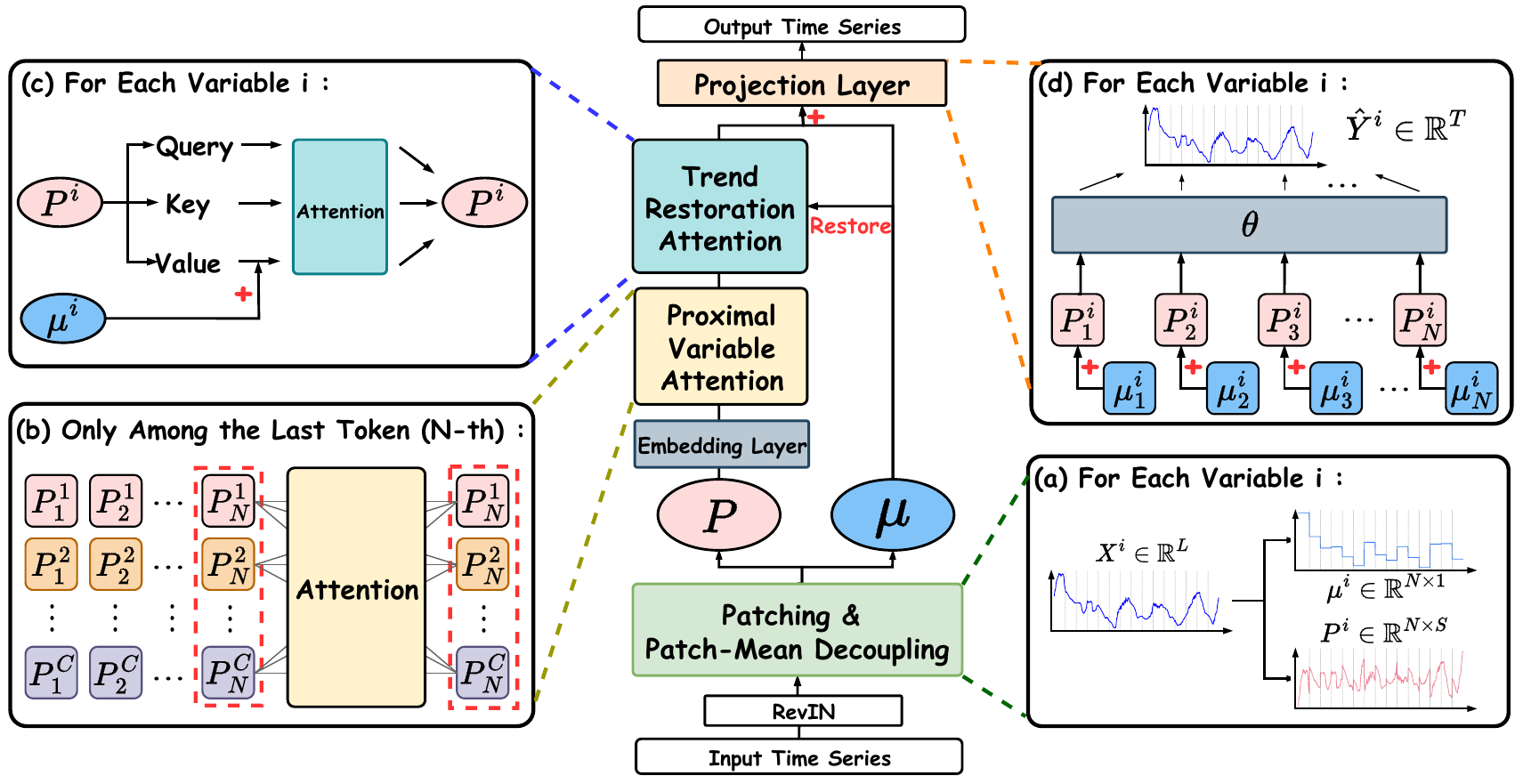}
\caption{Overview of the proposed PMDformer. The model comprises:
(a) \textbf{Patch-Mean Decoupling} module re-centers each patch and separates patches into trend and shape components;
(b) \textbf{Proximal Variable Attention} operates only on the most recent token to capture variable interactions which are most relevant for forecasting;
(c) \textbf{Trend Restoration Attention} restores long-range trends after value projections, restoring trend modeling;
(d) \textbf{Projection Linear} adds the trend back to model long-range trend information for stable and accurate predictions.}\label{fig:method}
\end{figure*}

Our proposed \textbf{PMDformer} architecture is a unified framework composed of four synergistic modules designed to explicitly decouple the long-term trend from the shape structure, selectively focus on the most relevant inter-variable dependencies, and ensure the accurate restoration of global dynamics for stable forecasting, as illustrated in Figure~\ref{fig:method}.
(a) \textbf{Patch-Mean Decoupling (PMD)}: This module partitions the input time series into non-overlapping patches and explicitly separates each patch into its long-term trend component and its residual shape component.
(b) \textbf{Proximal Variable Attention (PVA)}: To capture the most relevant cross-variable dependencies, the PVA module focuses its self-attention mechanism only on the $C$ tokens of the \textbf{last (proximal) patch}, modeling interactions across all variables.
(c) \textbf{Trend Restoration Attention (TRA)}: This module is designed to model the shape similarities across patches. Crucially, it then \textbf{restores} the long-range trend information into the value pathway, enabling to accurately capture and utilize the overall long-term trend.
(d) \textbf{Projection Layer}: This final layer combines the learned temporal representations with the reincorporated trend information through a fully connected projection to produce the final predictions.

\subsection{Model Architecture}
\paragraph{Patch-Mean Decoupling (PMD) \& Embedding.}
We first divide the input sequence $\mathbf{X}=\{x_t\in\mathbb{R}^{C}\}_{t=1}^{L}$ into $N$ non-overlapping patches of length $S$, where $N=\lfloor L/S \rfloor$. For variable $i\!\in\![C]$ and patch index $j\!\in\![N]$, the raw patch vector is
\begin{align}
\mathbf{P}_{j}^{i}=\big(x_{(j-1)S+1}^{i},\,x_{(j-1)S+2}^{i},\,\ldots,\,x_{jS}^{i}\big)\in\mathbb{R}^{S}.
\end{align}
We then compute its temporal mean and the corresponding mean-decoupled residual:
\begin{align}\label{eqa:mu}
\mu_{j}^{i}=\tfrac{1}{S}\sum_{k=1}^{S}x_{(j-1)S+k}^{i}, \qquad
\mathbf{r}_{j}^{i}=\mathbf{P}_{j}^{i}-\mu_{j}^{i}\,\mathbf{1}_{S},
\end{align}
where $\mathbf{1}_{S}$ is the $S$-dimensional all-ones vector. Each residual patch is then embedded into a $d$-dimensional representation through a shared linear projection. To encode location, we add learned positional embeddings to form the Transformer token:
\begin{align}
\mathbf{P}_{j}^{i}:=\mathbf{r}_{j}^{i}\,\mathbf{W}_E+\mathbf{b}_{E} + \mathbf{z}_{p_j}
\end{align}
where $\mathbf{W}_E\in\mathbb{R}^{S\times d}$, $\mathbf{b}_E\in\mathbb{R}^{d}$, and $\mathbf{z}_{p_j}\in\mathbb{R}^{d}$ denotes the positional embedding of patch $j$. By removing patch means before embedding, each patch is centered, which alleviates local inconsistencies across patches and variables so that attention mechanism can focus on shape similarities.

\paragraph{Proximal Variable Attention (PVA).}
Intuitively, accurate time series forecasting hinges on the immediate interactions between variables at the most recent time steps, as these dependencies are most indicative of near-term changes. Therefore, the PVA module is designed to concentrate its attention mechanism on the most \textbf{proximal} (i.e., most recent) tokens to model these critical cross-variable relationships.

Let $N$ be the index of the last (most recent) patch. We collect the most recent tokens of all $C$ variables, denoted as $\mathcal{P}_N=\{\mathbf{P}_{N}^{1},\ldots,\mathbf{P}_{N}^{C}\}$, where each token $\mathbf{P}_{N}^{i}\in\mathbb{R}^{d}$ is derived from the Patch-Mean Decoupling (PMD) embedding. The PVA then applies Multi-Head Self-Attention (MHSA) exclusively within the set $\mathcal{P}_N$ to effectively capture the cross-variable shape dependencies that are most relevant for forecasting. Following the attention mechanism, a Feed-Forward Network (FFN) is employed to enhance the non-linear feature representation:
\begin{align}
&\hat{\mathcal{P}}_N = \text{LayerNorm}(\text{MHSA}(\mathcal{P}_N) + \mathcal{P}_N), \\
&\mathcal{P}_N = \text{LayerNorm}(\text{FFN}(\hat{\mathcal{P}}_N) + \hat{\mathcal{P}}_N).
\end{align}
Tokens from the earlier historical patches, specifically those indexed $\{1,\ldots,N-1\}$, maintain their original representation derived from the PMD module. Following the PVA operation, the refined token set $\mathcal{P}_N$ is concatenated with these remaining historical tokens along the patch dimension to form the full sequence of shape embeddings, denoted as $\mathcal{P}\in\mathbb{R}^{C\times N \times d}$. This deliberate strategy of restricting cross-variable attention solely to the most proximal patch offers dual advantages: it \textbf{enhances model robustness} by avoiding spurious long-range couplings from historical noise, and it improves \textbf{computational efficiency} by reducing the complexity from $O(C^2 N)$ to $O(C^2)$.

\paragraph{Trend Restoration Attention (TRA).}
Following the refinement of the most proximal tokens by the PVA module, the TRA module aims to capture temporal shape similarities across all historical patches while preserving long-range trend information. This is achieved by applying a parameter-shared Transformer encoder (MHSA + FFN) along the patch axis for each variable independently.

In this design, the Query($\mathbf{Q}$) and Key($\mathbf{K}$) projections operate solely on the shape embeddings, ensuring that the resulting attention scores $\mathcal{A}$ emphasize precise inter-patch shape similarity. To counteract the potential loss of global dynamics inherent in shape-focused modeling, we explicitly \textbf{incorporate the per-patch mean} ($\mu^i$) into the \textbf{Value} ($\mathbf{V}$) pathway. {The additive reintegration is inspired by residual connections in ResNet~\citep{ResNet}}. Concretely, for the $i$-th variable’s patch sequence $\mathbf{P}^i\in\mathbb{R}^{N\times d}$, the computation is defined as:
\begin{align}
&\mathbf{Q}^i = \mathbf{P}^i \mathbf{W}_Q, \qquad \mathbf{K}^i = \mathbf{P}^i \mathbf{W}_K, \\
&\mathcal{A} = \text{Softmax}\Big(\frac{\mathbf{Q}^i (\mathbf{K}^i)^\top}{\sqrt{d}}\Big), \\
&\mathbf{V}^i = \mathbf{P}^i \mathbf{W}_V + \mu^i,
\end{align}
where $\mathbf{W}_Q, \mathbf{W}_K, \mathbf{W}_V$ are the projection matrices, and $\mu^i$ is the per-patch mean (Eq.~\ref{eqa:mu}), broadcast to match the dimensions of $\mathbf{P}^i \mathbf{W}_V$. This architectural separation allows the $\mathbf{Q}/\mathbf{K}$ pathway to model fine-grained local shape dependencies, while the $\mathbf{V}$ pathway ensures the preservation of the essential \textbf{global trend dynamics}. The resulting trend-integrated tokens are then refined through a Feed-Forward Network (FFN) to enhance the temporal representation learning.

\paragraph{Projection Layer.}
The temporal tokens produced by the TRA module are rich in shape dependencies but still require the final \textbf{restoration of the global trend information} for stable and accurate multi-step forecasting. This final step is essential to fully recover the original scale and long-term dynamics that were decoupled earlier. To achieve this, before generating the multi-step forecasts, we \textbf{re-incorporate} the per-patch trend means ($\mu^i$) into the refined shape embeddings:
\begin{align}
\hat{\mathbf{Y}}^i = (\mathbf{P}^i + \mu^i)\,\mathbf{W}_o + \mathbf{b}_o,\quad \hat{\mathbf{Y}}^i \in \mathbb{R}^{T}.
\end{align}
Here, $\mathbf{W}_o \in \mathbb{R}^{(N \times d) \times T}$ and $\mathbf{b}_o \in \mathbb{R}^{T}$ are the weight matrix and bias vector, respectively. The mean $\mu^i$ is implicitly broadcast to align with the dimensions of $\mathbf{P}^i$. This final step ensures the model’s predictions are well-calibrated with the long-range trend observed in the input series.

\subsection{Theoretical Analysis}

\textbf{Scale Bias Without Patch-Mean Decoupling (PMD)}
Consider embedding raw patches \(\tilde{\mathbf{x}} = \mathbf{r} + \mu \mathbf{1}\), where \(\mathbf{r}\) is the residual and \(\mu\) is the patch mean. The attention logit between tokens \((i, j)\) is given by:
\begin{align}
\tilde{z}_{ij} = \mathbf{q}_i^\top \mathbf{k}_j = \tilde{\mathbf{x}}_i^\top \mathbf{M} \tilde{\mathbf{x}}_j = \underbrace{\mu_i \mu_j \mathbf{1}^\top \mathbf{M} \mathbf{1}}_{\text{mean--mean}} + \underbrace{\mu_i \mathbf{1}^\top \mathbf{M} \mathbf{r}_j + \mu_j \mathbf{r}_i^\top \mathbf{M} \mathbf{1}}_{\text{cross}} + \underbrace{\mathbf{r}_i^\top \mathbf{M} \mathbf{r}_j}_{\text{residual similarity}},
\end{align}
where \(\mathbf{M} := \mathbf{W}_E^\top \mathbf{W}_Q^\top \mathbf{W}_K \mathbf{W}_E\) and \(\mathbf{1}\) is the all-ones vector. The first three terms depend on the means and can affect even dominate the residual similarity, inducing scale bias.

\paragraph{Proposition 1: Sufficient Condition for Level-Dominated Logits}

Let \(i\) be a fixed query. A sufficient condition for the mean-dependent part of \(\tilde{z}_{ij}\) to dominate the residual similarity uniformly over all keys \(j\) is:
\begin{align}
|\mu_i| |\mu_j| |\mathbf{1}^\top \mathbf{M} \mathbf{1}| \ge \|\mathbf{M}\|_2 \|\mathbf{r}_i\| \|\mathbf{r}_j\| + |\mu_i| \|\mathbf{M} \mathbf{1}\| \|\mathbf{r}_j\| + |\mu_j| \|\mathbf{M} \mathbf{1}\| \|\mathbf{r}_i\|,
\end{align}
where \(\|\cdot\|_2\) represents the spectral norm. This condition guarantees that the mean-dependent terms outweigh the residual term and cross terms, leading to scale-induced bias in attention. This confirms that attention is biased toward scale when the means are large, which motivates the need for patch-mean decoupling in our method.

\section{Experiment}
\subsection{Experiment Setup}
\paragraph{Datasets}
We conduct experiments on 8 widely-used and publicly available real-world datasets. These include: ECL\footnote{\begin{minipage}[t]{0.9\textwidth}\href{https://archive.ics.uci.edu/ml/datasets/ElectricityLoadDiagrams20112014}{https://archive.ics.uci.edu/ml/datasets/ElectricityLoadDiagrams20112014}\end{minipage}}, Traffic\footnote{\begin{minipage}[t]{0.9\textwidth}\href{https://pems.dot.ca.gov/}{https://pems.dot.ca.gov/}\end{minipage}}, Weather\footnote{\begin{minipage}[t]{0.9\textwidth}\href{https://www.bgc-jena.mpg.de/wetter/}{https://www.bgc-jena.mpg.de/wetter/}\end{minipage}}, Solar\footnote{\begin{minipage}[t]{0.9\textwidth}\href{http://www.nrel.gov/grid/solar-power-data.html}{http://www.nrel.gov/grid/solar-power-data.html}\end{minipage}}, ETTh1, ETTh2, ETTm1, and ETTm2\footnote{\begin{minipage}[t]{0.9\textwidth}\href{https://github.com/zhouhaoyi/ETDataset}{https://github.com/zhouhaoyi/ETDataset}\end{minipage}}. Following the experimental protocol established in prior work~\citep{TSLib, TFB, LargeTS}, we partition the datasets into training, validation, and test sets with the following ratios: 6:2:2 for the four ETT datasets and 7:1:2 for the remaining datasets. The detailed statistics of each dataset are summarized in Table~\ref{tab:datasets}.

\begin{table}[htbp]
  \caption{Characteristics of Long-term Time Series Datasets. This table summarizes key attributes of each dataset, including the application domain; the number of variables; total time points; data split ratios for training, validation, and testing and sampling interval.}
  \label{tab:datasets}
  \centering
  \begin{threeparttable}
    \renewcommand{\arraystretch}{1.15}
    \adjustbox{max width=\columnwidth}{
      \begin{tabular}{l|ccccc|c|c|c}
        \toprule
        \textbf{Domain} & \multicolumn{5}{c|}{\textbf{Electricity}} & \textbf{Weather} & \textbf{Energy} & {\textbf{Transportation}} \\
        \cmidrule(lr){2-6} \cmidrule(lr){7-7} \cmidrule(lr){8-8} \cmidrule(lr){9-9}
        \textbf{Dataset} & ETTh1 & ETTh2 & ETTm1 & ETTm2 & ECL & Weather & Solar-Energy & Traffic \\
        \midrule
        \textbf{Variables} & 7 & 7 & 7 & 7 & 321 & 21 & 137 & 862 \\
        \textbf{Time Points} & 14,400 & 14,400 & 57,600 & 57,600 & 26,304 & 52,696 & 52,560 & 17,544 \\
        \textbf{Split Ratio} & 6:2:2 & 6:2:2 & 6:2:2 & 6:2:2 & 7:1:2 & 7:1:2 & 7:1:2 & 7:1:2\\
        \textbf{Sampling} & 1 hr & 1 hr & 15 min & 15 min & 1 hr & 10 min & 10 min & 1 hr \\
        \bottomrule
      \end{tabular}
    }
  \end{threeparttable}
\end{table}

\paragraph{Baselines}
We compare PMDformer against 9 baselines, including state-of-the-art (SOTA) long-term forecasting models: TQNet~\citep{TQNet}, TimeBase~\citep{TimeBase}, SOFTS~\citep{SOFTS}, SparseTSF~\citep{SparseTSF}, ModernTCN~\citep{ModernTCN}, iTransformer~\citep{iTransformer}, TimeMixer~\citep{timemixer}, and PatchTST~\citep{PatchTST}.

\paragraph{Setups}
Consistent with prior research~\citep{TimeBase}, we use an input length $L$ of 720 and evaluate prediction lengths $T$ of \{96, 192, 336, 720\}. Results for TimeBase, SparseTSF, iTransformer, TimeMixer, and PatchTST are derived from the TimeBase study, while other outcomes are from our own experiments. All experiments are conducted using PyTorch~\citep{pytorch} on an NVIDIA A100 80GB GPU. The Adam optimizer~\citep{Adam} is employed, with learning rates chosen from \{2e-4, 5e-4, 1e-3, 1e-2\}. The number of patches $N$ is adjusted based on the requirements of each dataset.
\begin{table*}[t]
  \centering
  \caption{Comprehensive results for multivariable time series forecasting with a lookback window of 720 time steps. Performance metrics for TQNet~\citep{TQNet} and SOFTS~\citep{SOFTS} were obtained through our experiments, while results for other methods were sourced from TimeBase~\citep{TimeBase}. The best results are highlighted in \first{bold}, and the second-best are indicated with \second{underlining}.}\label{tab:main_results_paper}
  \begin{threeparttable}
  \begin{adjustbox}{max width=\textwidth,center}
  \begin{tabular}{c|c|cc|cc|cc|cc|cc|cc|cc|cc|cc}
    \toprule
    \multicolumn{2}{c}{Models} &
    \multicolumn{2}{c}{\textbf{PMDformer}} &
    \multicolumn{2}{c}{TQNet} &
    \multicolumn{2}{c}{TimeBase} &
    \multicolumn{2}{c}{SOFTS} &
    \multicolumn{2}{c}{SparseTSF} &
    \multicolumn{2}{c}{ModernTCN} &
    \multicolumn{2}{c}{iTransformer} &
    \multicolumn{2}{c}{TimeMixer} &
    \multicolumn{2}{c}{PatchTST} \\
    \cmidrule(lr){3-4}
    \cmidrule(lr){5-6}
    \cmidrule(lr){7-8}
    \cmidrule(lr){9-10}
    \cmidrule(lr){11-12}
    \cmidrule(lr){13-14}
    \cmidrule(lr){15-16}
    \cmidrule(lr){17-18}
    \cmidrule(lr){19-20}
    \multicolumn{2}{c}{Year} &
    \multicolumn{2}{c}{\textbf{ours}} &
    \multicolumn{2}{c}{2025} &
    \multicolumn{2}{c}{2025} &
    \multicolumn{2}{c}{2024} &
    \multicolumn{2}{c}{2024} &
    \multicolumn{2}{c}{2024} &
    \multicolumn{2}{c}{2024} &
    \multicolumn{2}{c}{2024} &
    \multicolumn{2}{c}{2023} \\
    \cmidrule(lr){3-4}
    \cmidrule(lr){5-6}
    \cmidrule(lr){7-8}
    \cmidrule(lr){9-10}
    \cmidrule(lr){11-12}
    \cmidrule(lr){13-14}
    \cmidrule(lr){15-16}
    \cmidrule(lr){17-18}
    \cmidrule(lr){19-20}
    \multicolumn{2}{c}{Metric} &
     {MSE} & {MAE} &
     {MSE} & {MAE} &
     {MSE} & {MAE} &
     {MSE} & {MAE} &
     {MSE} & {MAE} &
     {MSE} & {MAE} &
     {MSE} & {MAE} &
     {MSE} & {MAE} &
     {MSE} & {MAE} \\
    \toprule

    \multirow{5}{*}{\rotatebox{90}{ECL}} 
    &  {96} 
    & \first{0.122} & \first{0.214}         
    & {0.143}  & {0.244}   
    & {0.139}  & {0.231}   
    & {0.133}  & {0.229}   
    & {0.139}  & {0.239}     
    & \second{0.131}  & \second{0.227}
    & {0.135} & {0.233}                              
    & {0.142} & {0.247}
    & {0.141} & {0.240} \\

    & {192}
    & \first{0.140} & \first{0.231}       
    & {0.151} & {0.247}
    & {0.153} & {0.255}
    & {0.160} & {0.255}
    & {0.155} & {0.250}   
    & \second{0.145} & \second{0.241}
    & {0.155} & {0.253}  
    & {0.159} & {0.256}
    & {0.156} & {0.256}  \\                          

    & {336} 
    & \first{0.152} & \first{0.245}
    & {0.166} & \second{0.261}
    & {0.169} & {0.262}
    & {0.182} & {0.277}
    & {0.171} & {0.265} 
    & \second{0.162} & \second{0.261}
    & {0.169} & {0.267} 
    & {0.169} & {0.270}
    & {0.172} & {0.267}\\
    
    & {720}
    & \first{0.177} & \first{0.272}
    & {0.194} & {0.291} 
    & {0.207} & {0.294}
    & {0.224} & {0.310}
    & {0.208} & {0.300} 
    & \second{0.193} & \second{0.289}
    & {0.204} & {0.301} 
    & {0.209} & {0.313}
    & {0.208} & {0.299} \\ 

    \midrule

    & Avg & \first{0.148} & \first{0.241} & {0.164} & {0.261} & {0.167} & {0.258} & {0.175} & {0.268} & {0.180} & {0.264} & \second{0.158} & \second{0.255} & {0.166} & {0.264} & {0.170} & {0.272} & {0.169} & {0.266} \\
    \midrule

    \multirow{5}{*}{\rotatebox{90}{Traffic}} 
    & {96} 
    & \first{0.338} & \first{0.212}
    & {0.398} & {0.297}
    & {0.394} & {0.267}
    & \second{0.355} & {0.255}
    & {0.389} & {0.268} 
    & {0.382} & {0.267}
    & {0.374} & {0.273} 
    & {0.396} & {0.294}
    & {0.363} & \second{0.250}\\
    
    & {192} 
    & \second{0.367} & \first{0.227}
    & {0.397} & {0.277}
    & {0.403} & {0.271}
    & \first{0.365} & \second{0.258}
    & {0.399} & {0.272} 
    & {0.393} & {0.271}
    & {0.393} & {0.283} 
    & {0.404} & {0.295}
    & {0.382} & \second{0.258} \\
    
    & {336} 
    & \first{0.379} & \first{0.235}
    & {0.403} & {0.279}
    & {0.417} & {0.278}
    & \second{0.390} & {0.278}
    & {0.417} & {0.279}
    & {0.406} & {0.277}
    & {0.409} & {0.292}
    & {0.419} & {0.302}
    & {0.399} & \second{0.268} \\
    
    & {720} 
    & \first{0.426} & \first{0.262}
    & {0.448} & {0.304}
    & {0.456} & {0.298}
    & \second{0.429} & {0.294}
    & {0.449} & {0.299}
    & {0.452} & {0.305}
    & {0.450} & {0.314} 
    & {0.458} & {0.309}
    & {0.432} & \second{0.289}\\

    \midrule

    & Avg & \first{0.378} & \first{0.234} & 0.412 & 0.289 & 0.418 & 0.279 & \second{0.385} & 0.271 & {0.414} & {0.280} & 0.408 & 0.280 & 0.407 & 0.291 & 0.419 & 0.300 & 0.394 & \second{0.266}\\

    \midrule

    \multirow{5}{*}{\rotatebox{90}{Weather}} 
    & {96} 
    & \first{0.141} & \first{0.181}
    & \second{0.160} & {0.213}
    & {0.146} & \second{0.198} 
    & {0.165} & {0.219}
    & {0.174} & {0.231} 
    & {0.155} & {0.210} 
    & {0.159} & {0.212} 
    & {0.163} & {0.223}
    & {0.149} & {0.199}\\
    
    & {192} 
    & \first{0.185} & \first{0.226}
    & {0.212} & {0.261}
    & \first{0.185} & \second{0.241} 
    & {0.213} & {0.258}
    & {0.216} & {0.267} 
    & {0.205} & {0.254} 
    & {0.203} & {0.252}
    & {0.201} & {0.254}
    & {0.193} & {0.243}\\
    
    & {336} 
    & \first{0.236} & \first{0.274}
    & {0.260} & {0.299}
    & \first{0.236} & \second{0.281} 
    & {0.272} & {0.305}
    & {0.260} & \second{0.299} 
    & {0.255} & {0.290} 
    & {0.253} & {0.291}
    & {0.258} & {0.300}
    & {0.240} & {0.281} \\
    
    & {720} 
    & \first{0.305} & \first{0.323}
    & {0.328} & {0.343}
    & \second{0.309} & \second{0.331}
    & {0.380} & {0.371} 
    & {0.325} & {0.345} 
    & {0.317} & {0.336} 
    & {0.317} & {0.337} 
    & {0.329} & {0.348}
    & {0.312} & {0.334}\\

    \midrule

    & Avg & \first{0.217} & \first{0.251} & {0.240} & {0.279} & \second{0.219} & \second{0.263} & {0.258} & {0.288} & {0.244} & {0.286} & {0.233} & {0.273} & {0.233} & {0.273} & {0.238} & {0.281} & {0.224} & {0.264}\\
    \midrule

    \multirow{5}{*}{\rotatebox{90}{Solar}} 
    & {96} 
    & \first{0.160} & \first{0.193}
    & {0.181} & {0.242}
    & \second{0.179} & {0.248} 
    & {0.192} & \second{0.239} 
    & {0.205} & {0.241}
    & {0.196} & {0.258} 
    & {0.217} & {0.255} 
    & {0.232} & {0.271} 
    & {0.205} & \second{0.239} \\
    
    & {192} 
    & \first{0.178} & \first{0.211}
    & {0.203} & {0.261}
    & {0.213} & \second{0.252} 
    & \second{0.197} & {0.259}
    & {0.215} & {0.265} 
    & {0.224} & {0.280} 
    & {0.208} & {0.257}
    & {0.238} & {0.293}
    & {0.227} & {0.280} \\
    
    & {336} 
    & \first{0.190} & \first{0.218}
    & {0.219} & {0.272}
    & {0.222} & \second{0.261} 
    & \second{0.212} & {0.273}
    & {0.213} & {0.276} 
    & {0.240} & {0.288} 
    & {0.238} & {0.309}
    & {0.234} & {0.301}
    & {0.225} & {0.290} \\
    
    & {720} 
    & \first{0.196} & \first{0.221}
    & {0.231} & {0.281}
    & {0.235} & \second{0.264} 
    & \second{0.217} & {0.274}
    & {0.232} & {0.272} 
    & {0.246} & {0.299} 
    & {0.270} & {0.319} 
    & {0.273} & {0.319}
    & {0.249} & {0.291} \\

    \midrule

    & Avg & \first{0.181} & \first{0.211} & {0.209} & {0.264} & {0.216} & \second{0.254} & \second{0.201} & {0.264} & {0.216} & {0.264} & {0.227} & {0.281} & {0.233} & {0.285} & {0.244} & {0.296} & {0.227} & {0.275} \\
    \midrule
    
    \multirow{5}{*}{\rotatebox{90}{ETTh1}}            
    & {96} 
    & \second{0.356} & \second{0.388}
    & {0.379} & {0.404}
    & \first{0.349} & \first{0.384}
    & {0.389} & {0.417}
    & {0.362} & {0.389}
    & {0.380} & {0.405}
    & {0.389} & {0.421}
    & {0.410} & {0.441}
    & {0.377} & {0.408} \\

    & {192} 
    & \second{0.397} & {0.416}
    & {0.429} & {0.441}
    & \first{0.387} & \first{0.410}
    & {0.427} & \second{0.443}
    & {0.404} & \second{0.412}
    & {0.418} & {0.428}
    & {0.424} & {0.446}
    & {0.448} & {0.465}
    & {0.413} & {0.431} \\
    
    & {336} 
    & \second{0.420} & {0.432}
    & {0.454} & {0.455}
    & \first{0.408} & \first{0.418}
    & {0.446} & {0.458}
    & {0.435} & \second{0.428}
    & {0.453} & {0.450}
    & {0.456} & {0.469}
    & {0.482} & {0.490}
    & {0.436} & {0.446} \\

    & {720} 
    & \second{0.432} & {0.456}
    & {0.499} & {0.506}
    & {0.439} & \first{0.446}
    & {0.468} & {0.491}
    & \first{0.426} & \second{0.448}
    & {0.480} & {0.484}
    & {0.545} & {0.532}
    & {0.475} & {0.500}
    & {0.455} & {0.475} \\

    \midrule
    
    & Avg & \second{0.401} & {0.423} & {0.440} & {0.452} & \first{0.396} & \first{0.415} & {0.433} & {0.452} & {0.407} & \second{0.419} & {0.433} & {0.442} & {0.454} & {0.467} & {0.454} & {0.474} & {0.420} & {0.440} \\
    \midrule

    \multirow{5}{*}{\rotatebox{90}{ETTh2}}      
    & {96} 
    & \first{0.269} & \first{0.329}
    & {0.288} & {0.354}
    & {0.292} & {0.345}
    & {0.309} & {0.365}
    & {0.294} & {0.346}
    & \second{0.273} & \second{0.341}
    & {0.305} & {0.361}
    & {0.315} & {0.380}
    & {0.276} & {0.339} \\

    & {192} 
    & \first{0.333} & \first{0.373}
    & {0.377} & {0.403}
    & {0.339} & {0.387}
    & {0.378} & {0.405}
    & {0.340} & \second{0.377}
    & \second{0.337} & {0.385}
    & {0.405} & {0.421}
    & {0.383} & {0.415}
    & {0.342} & {0.385}\\

    & {336} 
    & \first{0.357} & \first{0.396}
    & {0.377} & {0.415}
    & {0.358} & {0.410}
    & {0.460} & {0.461}
    & \second{0.360} & \second{0.398}
    & {0.369} & {0.414}
    & {0.411} & {0.436}
    & {0.385} & {0.438}
    & {0.364} & {0.405}\\

    & {720} 
    & \second{0.390} & \second{0.429}
    & {0.424} & {0.452}
    & {0.400} & {0.448}
    & {0.441} & {0.467}
    & \first{0.383} & \first{0.425}
    & {0.408} & {0.448}
    & {0.448} & {0.470}
    & {0.432} & {0.471}
    & {0.395} & {0.434}\\

    \midrule

    & Avg & \first{0.337} & \first{0.382} & {0.367} & {0.406} & {0.347} & {0.398} & {0.397} & {0.425} & \second{0.344} & \second{0.387} & {0.347} & {0.397} & {0.392} & {0.422} & {0.379} & {0.426} & \second{0.344} & {0.391} \\ 
    \midrule
    
    \multirow{5}{*}{\rotatebox{90}{ETTm1}}            
    & {96} 
    & \first{0.279} & \first{0.328}
    & \second{0.296} & \second{0.349}
    & {0.311} & {0.351}
    & {0.303} & {0.361}
    & {0.314} & {0.359}
    & {0.313} & {0.357}
    & {0.315} & {0.369}
    & {0.332} & {0.384}
    & {0.298} & {0.352}\\
    
    & {192} 
    & \first{0.323} & \first{0.358}
    & {0.337} & {0.374}
    & {0.338} & \second{0.371}
    & \second{0.336} & {0.377}
    & {0.348} & {0.376}
    & {0.343} & {0.377}
    & {0.349} & {0.388}
    & {0.355} & {0.398}
    & \second{0.335} & {0.373}\\

    & {336} 
    & \first{0.361} & \first{0.383}
    & {0.369} & {0.393}
    & \second{0.364} & \second{0.386}
    & {0.384} & {0.407}
    & {0.368} & {0.386}
    & {0.372} & {0.393}
    & {0.381} & {0.409}
    & {0.386} & {0.416}
    & {0.366} & {0.394}\\
    
    & {720} 
    & {0.421} & {0.416}
    & {0.447} & {0.434}
    & \first{0.413} & \second{0.414}
    & {0.438} & {0.438}
    & {0.419} & \first{0.413}
    & \second{0.420} & {0.420}
    & {0.437} & {0.439}
    & {0.452} & {0.457}
    & {0.420} & {0.421}\\
    
    \midrule
    & Avg & \first{0.346} & \first{0.371} & {0.362} & {0.388} & {0.357} & \second{0.381} & {0.365} & {0.396} & {0.362} & {0.384} & {0.362} & {0.387} & {0.371} & {0.401} & {0.381} & {0.414} & \second{0.355} & {0.385} \\
    \midrule

    \multirow{5}{*}{\rotatebox{90}{ETTm2}}            
    & {96} 
    & \first{0.155} & \first{0.240}
    & {0.169} & \second{0.257}
    & {0.167} & {0.259}
    & {0.188} & {0.274}
    & {0.167} & {0.259}
    & {0.179} & {0.269}
    & {0.179} & {0.274}
    & {0.192} & {0.285}
    & \second{0.165} & {0.260}\\

    & {192} 
    & \first{0.213} & \first{0.282}
    & {0.231} & {0.299}
    & \second{0.219} & \second{0.297}
    & {0.256} & {0.317}
    & {0.219} & {0.297}
    & {0.243} & {0.312}
    & {0.239} & {0.314}
    & {0.253} & {0.329}
    & \second{0.219} & {0.298}\\
    
    & {336} 
    & \first{0.267} & \first{0.319}
    & {0.282} & {0.337}
    & {0.271} & {0.330}
    & {0.334} & {0.366}
    & {0.271} & {0.330}
    & {0.270} & \second{0.330}
    & {0.309} & {0.356}
    & {0.307} & {0.362}
    & \second{0.268} & {0.333}\\

    & {720} 
    & \first{0.347} & \first{0.373}
    & {0.371} & {0.398}
    & {0.353} & \second{0.380}
    & {0.392} & {0.406}
    & \second{0.353} & {0.380}
    & {0.362} & {0.393}
    & {0.387} & {0.407}
    & {0.380} & {0.412}
    & {0.352} & {0.386}\\

    \midrule

    & Avg & \first{0.246} & \first{0.304} & {0.263} & {0.323} & {0.253} & \second{0.317} & {0.293} & {0.341} & {0.253} & \second{0.317} & {0.264} & {0.326} & {0.279} & {0.338} & {0.283} & {0.347} & \second{0.251} & {0.319} \\ 
    \midrule 

    \multicolumn{2}{c}{\textbf{1$^{st}$ Count}} 
    & \first{32} & \first{33}
    & {0} & {0}
    & {7} & {4}
    & {1} & {0}
    & {2} & {3}
    & {0} & {0}
    & {0} & {0}
    & {0} & {0}
    & {0} & {0}\\

    \bottomrule
  \end{tabular}
  \end{adjustbox}
  \end{threeparttable}
\end{table*}

\subsection{Main Results}
 Table~\ref{tab:main_results_paper} summarizes the quantitative results for long-term time series forecasting across multiple prediction horizons and datasets. As shown, our proposed PMDformer achieves the lowest Mean Squared Error (MSE) and Mean Absolute Error (MAE) on 7 out of 8 real-world datasets, outperforming all baselines in the majority of cases. This success is directly tied to PMDformer's ability to overcome fundamental limitations in existing architectures.

Specifically, compared to the patch-based model TimeBase, PMDformer yields an average MSE reduction of 5.68\% and MAE reduction of 6.61\%. This improvement stems from our method's capacity to indentify meaningful shape similarities across patches, a capability that TimeBase's orthogonal patch selection inherently sacrifices to reduce redundancy. Moreover, against TQNet, PMDformer achieves an average MSE reduction of 8.62\% and MAE reduction of 9.96\%. TQNet's fixed periodic queries constrain its ability to handle diverse cycles, whereas PMDformer's adaptive proximal variable attention offers greater flexibility in modeling variables' shape similarities. 
Compared to the Transformer-based iTransformer, PMDformer delivers an average MSE reduction of 11.44\% and MAE reduction of 12.38\%. iTransformer captures dependencies among variable tokens embedded from the entire historical sequence, which can lead to overfitting on early, weakly relevant variable relationships that degrade future predictions. In contrast, our PVA module succeed to avoid this by focusing on the shape similarities of variables within the most nearest patch.

\begin{table*}[t]
  \centering
  \caption{Ablation study on PMD module. We assess different modules for patch-wise normalization, along with removing PMD module. Results are averaged across all prediction horizons.}\label{tab:ablation_pmd}
  \begin{adjustbox}{max width=\textwidth,center}
  \begin{tabular}{c|c|cc|cc|cc|cc|cc}
    \toprule
    \multirow{2}{*}{Design} & \multirow{2}{*}{Norm} & \multicolumn{2}{c|}{ETTh2} & \multicolumn{2}{c|}{ETTm1} & 
    \multicolumn{2}{c|}{Weather} & 
    \multicolumn{2}{c|}{Traffic} & \multicolumn{2}{c}{Solar} \\
    \cmidrule(lr){3-4} \cmidrule(lr){5-6} \cmidrule(lr){7-8} \cmidrule(lr){9-10}\cmidrule(lr){11-12}
           &               & MSE   & MAE   & MSE   & MAE   & MSE   & MAE   & MSE   & MAE   & MSE   & MAE\\
    \midrule
    PMDformer & PMD & \first{0.337} & \first{0.382} & \first{0.346} & \first{0.371} & \first{0.217} & \first{0.251} & \first{0.378} & 
    \first{0.234} & \first{0.181} & \first{0.211} \\
    \midrule
    \multirow{3}{*}{Replace} & w/ stdev & 0.354 & 0.392 & 0.347 & 0.370 & 0.218 & 0.252 & 0.396 & 0.259 & 0.205 & 0.221 \\
            & SAN    & 0.360 & 0.403 & 0.353 & 0.380 & 0.225 & 0.275 & 0.392 & 0.273 & 0.182 & 0.235 \\
            & \ding{55}    & 0.359 & 0.394 & 0.347 & 0.370 & 0.223 & 0.260 & 0.397 & 0.258 & 0.199 & 0.212 \\
    \bottomrule
  \end{tabular}
    \end{adjustbox}
\end{table*}

\begin{table*}[t]
  \centering
  \caption{Ablation studies on TRA and PVA modules in PMDformer: Performance impacts of replacements, removals, and order swaps across ETTh2, ETTm1, Traffic, and Solar datasets. Results are averaged across all prediction horizons.}\label{tab:ablation_TVA_TRA}
  \begin{adjustbox}{max width=\textwidth,center}
  \begin{tabular}{c|c|c|cc|cc|cc|cc}
    \toprule
    \multirow{2}{*}{Design} & \multirow{2}{*}{TRA} & \multirow{2}{*}{PVA} & \multicolumn{2}{c|}{ETTh2} & \multicolumn{2}{c|}{ETTm1} & \multicolumn{2}{c|}{Traffic} & \multicolumn{2}{c}{Solar} \\
    \cmidrule(lr){4-5} \cmidrule(lr){6-7} \cmidrule(lr){8-9} \cmidrule(lr){10-11}
           &          &          & MSE   & MAE   & MSE   & MAE   & MSE   & MAE   & MSE   & MAE   \\
    \midrule
    PMDformer & \ding{51} & Last Token & \first{0.337} & \first{0.382} & \first{0.346} & \first{0.371} & \first{0.378} & \first{0.234} & \first{0.181} & \first{0.211} \\
    \midrule
    \multirow{3}{*}{Replace} & \ding{51} & All Token  & 0.340 & 0.384 & 0.354 & 0.375 & 0.380 & 0.239 & 0.186 & 0.214 \\
     & Self-attention & Last Token  & 0.345 & 0.386 & 0.352 & 0.372 & 0.388 & 0.251 & 0.196 & 0.217 \\
     & \multicolumn{2}{c|}{Swap Order $\leftrightharpoons$} & 0.342 & 0.385 & 0.350 & 0.372 & 0.379 & 0.235 & 0.188 & 0.216 \\
    \midrule
    \multirow{3}{*}{w/o}     & \ding{55} & Last Token        & 0.344 & 0.381 & 0.347 & 0.372 & 0.410 & 0.270 & 0.215 & 0.226 \\
    & \ding{51} & \ding{55}        & 0.340 & 0.383 & 0.347 & 0.371 & 0.386 & 0.240 & 0.194 & 0.214 \\
    & \ding{55} & \ding{55}       & 0.346 & 0.384 & 0.351 & 0.372 & 0.426 & 0.288 & 0.222 & 0.230 \\
    \bottomrule
  \end{tabular}
  \end{adjustbox}
\end{table*}

\paragraph{PMD Module Analysis.}
We assessed the effectiveness of the PMD module through extensive ablations conducted on five non-stationary benchmarks: ETTh2, ETTm1, Weather, Traffic, and Solar~\citep{ab_PMD_1,ab_PMD_2}. Using a fixed input length of 720, we tested the model's performance across various prediction horizons (96, 192, 336, and 720) against several patch-wise normalization variants: (i) mean–variance standardization ('w/ stdev'), (ii) utilizing the Scale-Adaptive Normalization (SAN)~\citep{SAN} method, and (iii) removing the PMD module entirely. As presented in Table~\ref{tab:ablation_pmd}, the PMDformer consistently achieves superior accuracy across all datasets. We attribute this advantage to the PMD module’s per-patch centering mechanism, which effectively preserves crucial intra-patch shape information. This preservation allows the Transformer architecture to specifically concentrate its attention on modeling \textbf{shape similarity}. Furthermore, by explicitly injecting the patch mean as a separated trend component into the Transformer pathway, PMDformer is uniquely positioned to accurately capture and model long-term trends. In stark contrast, SAN explicitly decouples the scale and residual components for independent prediction. Since global scale estimation is inherently unstable in highly non-stationary series, this rigid decoupling undermines the essential joint modeling of scale–shape interactions, consequently leading to overfitting and weaker generalization capabilities.

\paragraph{TRA \& PVA Analysis.}
To assess the effectiveness of the TRA and PVA modules, we conducted ablation studies on the ETTh2, ETTm1, Traffic, and Solar datasets. For the TRA module, we tested two alternatives: replacing it with standard self-attention or removing it entirely. For the PVA module, we either modified it to compute variable-wise shape similarity across \textit{all} patches or removed the module completely. Additionally, we investigated a structural variant that swaps the sequential order of the two modules. The experimental outcomes are summarized in Table~\ref{tab:ablation_TVA_TRA}.

The results unequivocally show that $\text{PMDformer}$ consistently outperforms all ablated variants across every dataset and configuration. When TRA is replaced with standard self-attention, performance degrades significantly because the crucial long-term trend information is neglected. Similarly, when PVA is forced to compute variable-wise shape similarity across all historical patches, performance decreases. This confirms our hypothesis that early variable relationships are often only weakly or spuriously correlated with the predictive sequences, justifying PVA's proximal focus. Furthermore, removing both TRA and PVA results in the largest performance drop observed, emphatically highlighting the dual importance of TRA in modeling temporal patch shapes and long-range trends, and PVA in capturing relevant variable-wise shape similarity. Finally, swapping the original order of TRA and PVA also causes notable performance degradation. When TRA is applied first, it compresses patch information too early, making it harder for the subsequent variable modeling to identify meaningful cross-variable dependencies.

\subsection{Parameter Sensitivity Analysis}
\begin{figure*}[t]   
    \centering
    \includegraphics[width=\textwidth]{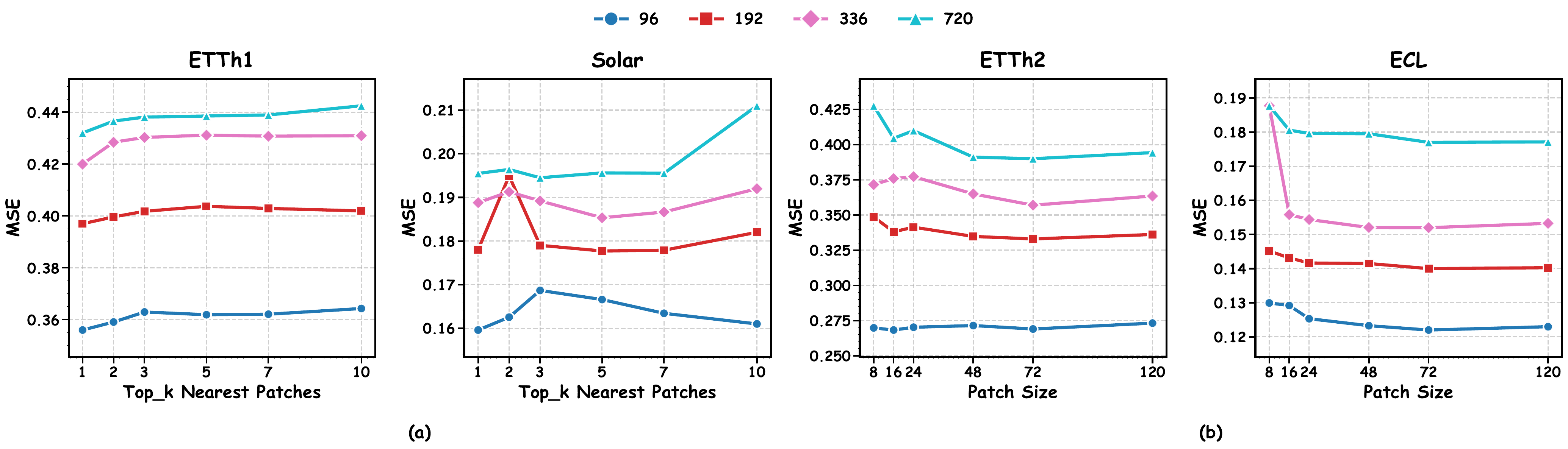}
\caption{Parameter Sensitivity Analysis. (a) Selection of the number of $k$ nearest patches to the prediction sequence for capturing inter-variable dependencies on these patches. Superior and more stable performance is achieved when $k=1$. (b) Different patch sizes are used to partition the input sequence, and a moderate patch size yields the optimal choice.}\label{fig:para_sensity} 
\end{figure*}

\paragraph{Patch Count for Cross-Variable Modeling.}
We evaluate the impact of capturing variable patterns within different numbers of patches, where $k \in \{1,2,3,5,7,10\}$. For each setting, the $k$ nearest patches to future sequences are selected to capture the shape similarity of variable, thereby further validating the effectiveness of PVA. Experiments are conducted on the ETTh1 and Solar datasets. As shown in Figure~\ref{fig:para_sensity} (a), the mean squared error (MSE) exhibits an overall upward trend as $k$ gradually increases on the ETTh1 dataset. On the Solar dataset, this increase is more pronounced when predicting 192, 336, or 720 steps ahead, because future sequences are more weakly correlated with early variable relationships. Moreover, the MSE curves show some fluctuations, indicating that different values of $k$ may lead to more significant differences in prediction performance. In contrast, across all four prediction horizons, using $k=1$ yields more stable performance compared with larger $k$. This is because the nearest patch is typically more closely aligned with the target sequence to be predicted, making it more beneficial for accurate modeling.

\paragraph{Patch Size.}
Different patch sizes lead to varying degrees of distinction among patches. To investigate this, we evaluate multiple patch sizes $\{8, 16, 24, 48, 72, 120\}$ on the ETTh2 and ECL datasets. As shown in {Figure~\ref{fig:para_sensity} (b)}, both overly small and overly large patch sizes fail to deliver optimal performance. This is because excessively small patches provide insufficient shape information to distinguish similarity, making it difficult for the attention mechanism to capture underlying temporal dependencies or genuine variable correlations. Conversely, overly large patches reduce the number of tokens, thereby limiting the model’s ability to capture long-range dependencies. Based on these observations, we find that moderate patch sizes, particularly within $\{24, 48, 72\}$, achieve a better trade-off and yield more robust performance.

\section{Conclusion}
In this paper, we tackle challenges in long-term time series forecasting by emphasizing true shape similarities hidden by scale variations in non-stationary data. Our patch-mean decoupling (PMD) separates trends from residual shapes while preserving amplitudes, enabling shape-focused attention across patches and variables. Integrated with proximal variable attention (PVA) for recent inter-variable dependencies and trend restoration attention (TRA) for global trend reintegration.
Experiments on LTSF benchmarks show PMDformer surpasses state-of-the-art baselines in accuracy and stability, underscoring the value of shape-centric Transformer designs. Future directions include scaling to higher-dimensional multivariate data and multimodal integrations for applications in energy, finance, and traffic.

\section{Acknowledgements}
This work was supported by the Major Science and Technology Special Project of the Sichuan Provincial Department of Science and Technology (Grant No. 2024ZDZX0002), the Sichuan Provincial Innovation Group Project (Grant No. 2024NSFTD0054), Fundamental Research Funds for the Central Universities (JBK202511081), the Blockchain Research Center of China, the Natural Science Foundation of China (Grant No. 62502397), the National Natural Science Foundation of China (Grant No. 72471197), and the Sichuan Provincial Philosophy and Social Science Fund (Grant No. SCJJ25ND091).



\bibliography{iclr2026/iclr2026_conference}

@inproceedings{timemixer,
  title={TimeMixer: Decomposable Multiscale Mixing for Time Series Forecasting},
  author={Wang, Shiyu and Wu, Haixu and Shi, Xiaoming and Hu, Tengge and Luo, Huakun and Ma, Lintao and Zhang, James Y and ZHOU, JUN},
  booktitle={International Conference on Learning Representations (ICLR)},
  year={2024}
}

@inproceedings{TQNet,
  title={Temporal Query Network for Efficient Multivariate Time Series Forecasting}, 
  author={Lin, Shengsheng and Chen, Haojun and Wu, Haijie and Qiu, Chunyun and Lin, Weiwei},
  booktitle={Forty-second International Conference on Machine Learning},
  year={2025}
}

@article{SOFTS,
  title={SOFTS: Efficient Multivariate Time Series Forecasting with Series-Core Fusion},
  author={Han, Lu and Chen, Xu-Yang and Ye, Han-Jia and Zhan, De-Chuan},
  journal={arXiv preprint arXiv:2404.14197},
  year={2024}
}

@inproceedings{TimeBase,
  title     = {TimeBase: The Power of Minimalism in Long-term Time Series Forecasting},
  author    = {Qihe Huang and Zhengyang Zhou and Kuo Yang and Zhongchao Yi and Xu Wang and Yang Wang},
  booktitle = {Proceedings of the Forty-Second International Conference on Machine Learning (ICML)},
  year      = {2025}
}

@inproceedings{SparseTSF,
  title={SparseTSF: Modeling Long-term Time Series Forecasting with 1k Parameters},
  author={Lin, Shengsheng and Lin, Weiwei and Wu, Wentai and Chen, Haojun and Yang, Junjie},
  booktitle={Forty-first International Conference on Machine Learning},
  year={2024}
}

@article{iTransformer,
  title={iTransformer: Inverted Transformers Are Effective for Time Series Forecasting},
  author={Liu, Yong and Hu, Tengge and Zhang, Haoran and Wu, Haixu and Wang, Shiyu and Ma, Lintao and Long, Mingsheng},
  journal={ICLR},
  year={2024}
}

@article{PatchTST,
  title={A Time Series is Worth 64 Words: Long-term Forecasting with Transformers},
  author={Nie, Yuqi and Nguyen, Nam H and Sinthong, Phanwadee and Kalagnanam, Jayant},
  journal={ICLR},
  year={2023}
}

@article{finance_1,
  title={FinTSB: A Comprehensive and Practical Benchmark for Financial Time Series Forecasting}, 
  author={Yifan Hu and Yuante Li and Peiyuan Liu and Yuxia Zhu and Naiqi Li and Tao Dai and Shu-tao Xia and Dawei Cheng and Changjun Jiang},
  journal={arXiv preprint arXiv:2502.18834},
  year={2025},
}

@book{energy_1,
author = {Box, George Edward Pelham and Jenkins, Gwilym},
title = {Time Series Analysis, Forecasting and Control},
year = {1990},
publisher = {Holden-Day, Inc.},
}

@article{traffic_1, 
    title={Attention Based Spatial-Temporal Graph Convolutional Networks for Traffic Flow Forecasting},
    journal={Proceedings of the AAAI Conference on Artificial Intelligence}, 
    author={Guo, Shengnan and Lin, Youfang and Feng, Ning and Song, Chao and Wan, Huaiyu}, 
    year={2019} 
}

@inproceedings{traffic_2,
title={Fourier{GNN}: Rethinking Multivariate Time Series Forecasting from a Pure Graph Perspective},
author={Kun Yi and Qi Zhang and Wei Fan and Hui He and Liang Hu and Pengyang Wang and Ning An and Longbing Cao and Zhendong Niu},
booktitle={Thirty-seventh Conference on Neural Information Processing Systems},
year={2023}
}

@article{transformer,
  title={Attention is all you need},
  author={Vaswani, Ashish and Shazeer, Noam and Parmar, Niki and Uszkoreit, Jakob and Jones, Llion and Gomez, Aidan N and Kaiser, {\L}ukasz and Polosukhin, Illia},
  journal={Advances in neural information processing systems},
  year={2017}
}

@article{nonstationary_1,
  title={Non-stationary transformers: Exploring the stationarity in time series forecasting},
  author={Liu, Yong and Wu, Haixu and Wang, Jianmin and Long, Mingsheng},
  journal={Advances in neural information processing systems},
  volume={35},
  pages={9881--9893},
  year={2022}
}

@inproceedings{DishTS,
  title={Dish-ts: a general paradigm for alleviating distribution shift in time series forecasting},
  author={Fan, Wei and Wang, Pengyang and Wang, Dongkun and Wang, Dongjie and Zhou, Yuanchun and Fu, Yanjie},
  booktitle={Proceedings of the AAAI conference on artificial intelligence},
  year={2023}
}

@article{fouriergnn,
  title={FourierGNN: Rethinking multivariate time series forecasting from a pure graph perspective},
  author={Yi, Kun and Zhang, Qi and Fan, Wei and He, Hui and Hu, Liang and Wang, Pengyang and An, Ning and Cao, Longbing and Niu, Zhendong},
  journal={Advances in neural information processing systems},
  volume={36},
  pages={69638--69660},
  year={2023}
}

@inproceedings{Crossformer,
  title={Crossformer: Transformer utilizing cross-dimension dependency for multivariate time series forecasting},
  author={Zhang, Yunhao and Yan, Junchi},
  booktitle={The eleventh international conference on learning representations},
  year={2023}
}

@article{Pathformer,
  title={Pathformer: Multi-scale transformers with adaptive pathways for time series forecasting},
  author={Chen, Peng and Zhang, Yingying and Cheng, Yunyao and Shu, Yang and Wang, Yihang and Wen, Qingsong and Yang, Bin and Guo, Chenjuan},
  journal={arXiv preprint arXiv:2402.05956},
  year={2024}
}

@inproceedings{SAN,
  title={Adaptive Normalization for Non-stationary Time Series Forecasting: A Temporal Slice Perspective},
  author={Liu, Zhiding and Cheng, Mingyue and Li, Zhi and Huang, Zhenya and Liu, Qi and Xie, Yanhu and Chen, Enhong},
  booktitle={Thirty-seventh Conference on Neural Information Processing Systems},
  year={2023}
}

@inproceedings{TSMixer,
author = {Ekambaram, Vijay and Jati, Arindam and Nguyen, Nam and Sinthong, Phanwadee and Kalagnanam, Jayant},
title = {TSMixer: Lightweight MLP-Mixer Model for Multivariate Time Series Forecasting},
year = {2023},
booktitle = {Proceedings of the 29th ACM SIGKDD Conference on Knowledge Discovery and Data Mining},
}

@inproceedings{ModernTCN,
  title={Moderntcn: A modern pure convolution structure for general time series analysis},
  author={Luo, Donghao and Wang, Xue},
  booktitle={The twelfth international conference on learning representations},
  year={2024}
}

@article{timexer,
  title={Timexer: Empowering transformers for time series forecasting with exogenous variables},
  author={Wang, Yuxuan and Wu, Haixu and Dong, Jiaxiang and Liu, Yong and Qiu, Yunzhong and Zhang, Haoran and Wang, Jianmin and Long, Mingsheng},
  journal={Advances in Neural Information Processing Systems},
  year={2024}
}

@inproceedings{Informer,
  title={Informer: Beyond efficient transformer for long sequence time-series forecasting},
  author={Zhou, Haoyi and Zhang, Shanghang and Peng, Jieqi and Zhang, Shuai and Li, Jianxin and Xiong, Hui and Zhang, Wancai},
  booktitle={AAAI},
  year={2021}
}

@inproceedings{Pyraformer,
  title={Pyraformer: Low-complexity pyramidal attention for long-range time series modeling and forecasting},
  author={Liu, Shizhan and Yu, Hang and Liao, Cong and Li, Jianguo and Lin, Weiyao and Liu, Alex X and Dustdar, Schahram},
  booktitle={ICLR},
  year={2022}
}

@article{timer,
  title={Timer: Generative pre-trained transformers are large time series models},
  author={Liu, Yong and Zhang, Haoran and Li, Chenyu and Huang, Xiangdong and Wang, Jianmin and Long, Mingsheng},
  journal={arXiv preprint arXiv:2402.02368},
  year={2024}
}

@inproceedings{Moirai,
  title={Unified training of universal time series forecasting transformers},
  author={Woo, Gerald and Liu, Chenghao and Kumar, Akshat and Xiong, Caiming and Savarese, Silvio and Sahoo, Doyen},
  year={2024},
  booktitle={ICML}
}

@inproceedings{TimesFM,
  title={A decoder-only foundation model for time-series forecasting},
  author={Das, Abhimanyu and Kong, Weihao and Sen, Rajat and Zhou, Yichen},
  booktitle={Forty-first International Conference on Machine Learning},
  year={2024}
}

@article{TFB,
  title={Tfb: Towards comprehensive and fair benchmarking of time series forecasting methods},
  author={Qiu, Xiangfei and Hu, Jilin and Zhou, Lekui and Wu, Xingjian and Du, Junyang and Zhang, Buang and Guo, Chenjuan and Zhou, Aoying and Jensen, Christian S and Sheng, Zhenli and others},
  journal={arXiv preprint arXiv:2403.20150},
  year={2024}
}

@article{TSLib,
  title={Deep time series models: A comprehensive survey and benchmark},
  author={Wang, Yuxuan and Wu, Haixu and Dong, Jiaxiang and Liu, Yong and Long, Mingsheng and Wang, Jianmin},
  journal={arXiv preprint arXiv:2407.13278},
  year={2024}
}

@inproceedings{LargeTS,
  title={LargeST: A Benchmark Dataset for Large-Scale Traffic Forecasting},
  author={Liu, Xu and Xia, Yutong and Liang, Yuxuan and Hu, Junfeng and Wang, Yiwei and Bai, Lei and Huang, Chao and Liu, Zhenguang and Hooi, Bryan and Zimmermann, Roger},
  booktitle={Advances in Neural Information Processing Systems},
  year={2023}
}

@article{VIT,
  title={An image is worth 16x16 words: Transformers for image recognition at scale},
  author={Dosovitskiy, Alexey and Beyer, Lucas and Kolesnikov, Alexander and Weissenborn, Dirk and Zhai, Xiaohua and Unterthiner, Thomas and Dehghani, Mostafa and Minderer, Matthias and Heigold, Georg and Gelly, Sylvain and others},
  journal={arXiv preprint arXiv:2010.11929},
  year={2020}
}

@article{Adam,
  title={Adam: A method for stochastic optimization},
  author={Kingma, Diederik P},
  journal={arXiv preprint arXiv:1412.6980},
  year={2014}
}

@article{pytorch,
  title={Pytorch: An imperative style, high-performance deep learning library},
  author={Paszke, Adam and Gross, Sam and Massa, Francisco and Lerer, Adam and Bradbury, James and Chanan, Gregory and Killeen, Trevor and Lin, Zeming and Gimelshein, Natalia and Antiga, Luca and others},
  journal={Advances in neural information processing systems},
  volume={32},
  year={2019}
}

@inproceedings{shape_1,
  title={Learning time-series shapelets},
  author={Grabocka, Josif and Schilling, Nicolas and Wistuba, Martin and Schmidt-Thieme, Lars},
  booktitle={Proceedings of the 20th ACM SIGKDD international conference on Knowledge discovery and data mining},
  pages={392--401},
  year={2014}
}

@inproceedings{shape_2,
  title={Towards transparent time series forecasting},
  author={Kacprzyk, Krzysztof and Liu, Tennison and van der Schaar, Mihaela},
  booktitle={The Twelfth International Conference on Learning Representations},
  year={2024}
}

@article{shape_3,
  title={Shape analysis for time series},
  author={Germain, Thibaut and Gruffaz, Samuel and Truong, Charles and Durmus, Alain and Oudre, Laurent},
  journal={Advances in neural information processing systems},
  volume={37},
  pages={95607--95638},
  year={2024}
}

@article{ab_PMD_1,
  title={Onenet: Enhancing time series forecasting models under concept drift by online ensembling},
  author={Wen, Qingsong and Chen, Weiqi and Sun, Liang and Zhang, Zhang and Wang, Liang and Jin, Rong and Tan, Tieniu and others},
  journal={Advances in Neural Information Processing Systems},
  volume={36},
  pages={69949--69980},
  year={2023}
}

@inproceedings{ab_PMD_2,
  title={Battling the non-stationarity in time series forecasting via test-time adaptation},
  author={Kim, HyunGi and Kim, Siwon and Mok, Jisoo and Yoon, Sungroh},
  booktitle={AAAI},
  year={2025}
}

@inproceedings{SIN,
  title={SIN: Selective and interpretable normalization for long-term time series forecasting},
  author={Han, Lu and Ye, Han-Jia and Zhan, De-Chuan},
  booktitle={Forty-first International Conference on Machine Learning},
  year={2024}
}

@book{shape_4,
  title={Time series analysis},
  author={Hamilton, James D},
  year={2020},
  publisher={Princeton university press}
}

@article{RLinear,
  title={Revisiting long-term time series forecasting: An investigation on linear mapping},
  author={Li, Zhe and Qi, Shiyi and Li, Yiduo and Xu, Zenglin},
  journal={arXiv preprint arXiv:2305.10721},
  year={2023}
}

@article{autoformer,
  title={Autoformer: Decomposition transformers with auto-correlation for long-term series forecasting},
  author={Wu, Haixu and Xu, Jiehui and Wang, Jianmin and Long, Mingsheng},
  journal={Advances in neural information processing systems},
  volume={34},
  pages={22419--22430},
  year={2021}
}

@inproceedings{FEDformer,
  title={Fedformer: Frequency enhanced decomposed transformer for long-term series forecasting},
  author={Zhou, Tian and Ma, Ziqing and Wen, Qingsong and Wang, Xue and Sun, Liang and Jin, Rong},
  booktitle={International conference on machine learning},
  pages={27268--27286},
  year={2022},
  organization={PMLR}
}

@article{DLinear,
  title={Are Transformers Effective for Time Series Forecasting?},
  author={Ailing Zeng and Muxi Chen and Lei Zhang and Qiang Xu},
  journal={AAAI},
  year={2023}
}

@article{CrossGNN,
  title={CrossGNN: Confronting Noisy Multivariate Time Series Via Cross Interaction Refinement},
  author={Huang, Qihe and Shen, Lei and Zhang, Ruixin and Ding, Shouhong and Wang, Binwu and Zhou, Zhengyang and Wang, Yang},
  journal={NeurIPS},
  year={2023},
}

@inproceedings{micn,
  title={MICN: Multi-scale Local and Global Context Modeling for Long-term Series Forecasting},
  author={Huiqiang Wang and Jian Peng and Feihu Huang and Jince Wang and Junhui Chen and Yifei Xiao},
  booktitle={ICLR},
  year={2023}
}

@article{TSLANet,
  title={Tslanet: Rethinking transformers for time series representation learning},
  author={Eldele, Emadeldeen and Ragab, Mohamed and Chen, Zhenghua and Wu, Min and Li, Xiaoli},
  journal={arXiv preprint arXiv:2404.08472},
  year={2024}
}

@article{PatchMixer,
  title={Patchmixer: A patch-mixing architecture for long-term time series forecasting},
  author={Gong, Zeying and Tang, Yujin and Liang, Junwei},
  journal={arXiv preprint arXiv:2310.00655},
  year={2023}
}

@inproceedings{S2LLM,
  title={S$^2$IP-LLM: Semantic space informed prompt learning with LLM for time series forecasting},
  author={Pan, Zijie and Jiang, Yushan and Garg, Sahil and Schneider, Anderson and Nevmyvaka, Yuriy and Song, Dongjin},
  booktitle={Forty-first International Conference on Machine Learning},
  year={2024}
}

@article{time-llm,
  title={Time-llm: Time series forecasting by reprogramming large language models},
  author={Jin, Ming and Wang, Shiyu and Ma, Lintao and Chu, Zhixuan and Zhang, James Y and Shi, Xiaoming and Chen, Pin-Yu and Liang, Yuxuan and Li, Yuan-Fang and Pan, Shirui and others},
  journal={arXiv preprint arXiv:2310.01728},
  year={2023}
}

@inproceedings{RevIN,
  title={Reversible instance normalization for accurate time-series forecasting against distribution shift},
  author={Kim, Taesung and Kim, Jinhee and Tae, Yunwon and Park, Cheonbok and Choi, Jang-Ho and Choo, Jaegul},
  booktitle={International conference on learning representations},
  year={2021}
}

@inproceedings{ResNet,
  author       = {Kaiming He and
                  Xiangyu Zhang and
                  Shaoqing Ren and
                  Jian Sun},
  title        = {Deep Residual Learning for Image Recognition},
  booktitle    = {CVPR},
  year         = {2016}
}

@article{TimeCNN,
  title={TimeCNN: Refining Cross-Variable Interaction on Time Point for Time Series Forecasting},
  author={Hu, Ao and Wen, Liangjian and Dai, Yong and Qi, Shiyi and Wang, Jun and Chen, Zhi and Zhou, Xun and Wang, Dongkai and Xu, Zenglin and Duan, Jiang},
  journal={Neural Networks},
  year={2025}
}

@article{FDNet,
  title={FDNet: High-Frequency Disentanglement Network with Information-Theoretic Guidance for Multivariate Time Series Forecasting},
  author={Hu, Ao and Wen, Liangjian and Duan, Jiang and Dai, Yong and Wang, Dongkai and Huang, Shudong and Wang, Jun and Xu, Zenglin},
  journal={Pattern Recognition},
  year={2025}
}
\bibliographystyle{iclr2026/iclr2026_conference}

\appendix
\clearpage
\section{Appendix}
\subsection{Efficiency Analysis}
\begin{figure*}[htbp]   
    \centering
    \includegraphics[scale=0.3]{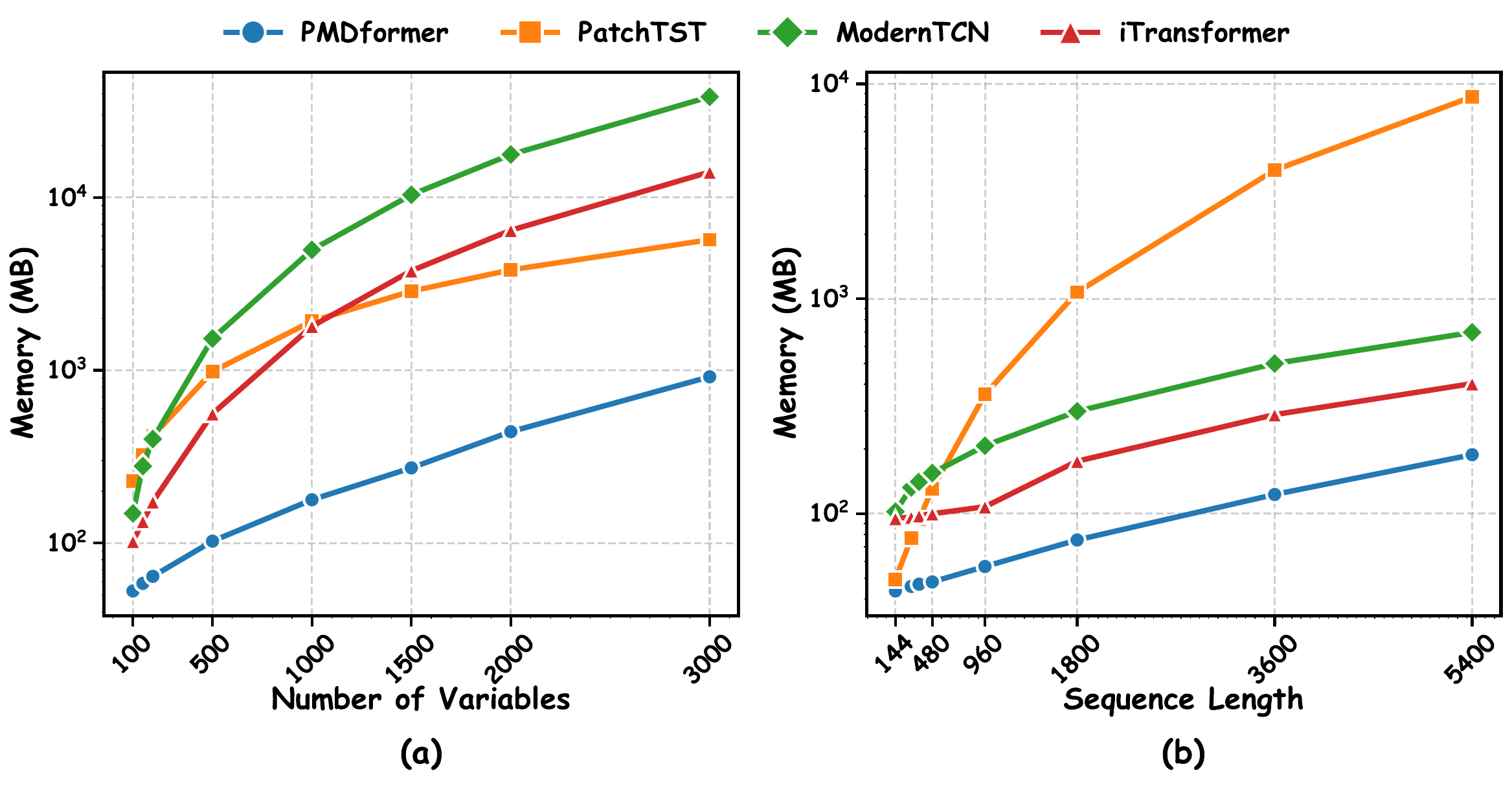}
\caption{{(a) Comparison of memory usage with varying number of variables $C$. (b) Comparison of memory usage with varying input sequence length $L$. PMDformer consistently requires the lowest memory.}}\label{fig:memory_comparison} 
\end{figure*}

{To evaluate the efficiency of our model in handling complex tasks, we conduct experiments under two settings: varying the number of variables and varying the input length. In the first setting, we fix the input length at 720 and change the number of variables; in the second setting, we fix the number of variables at 100 and test PMDformer with different input lengths. The batch size is set to 1 in all experiments. The results are shown in Figure~\ref{fig:memory_comparison}. Under both settings, compared with recent popular models such as PatchTST~\citep{PatchTST}, iTransformer~\citep{iTransformer}, and ModernTCN~\citep{ModernTCN}, PMDformer requires significantly less GPU memory, thereby reducing the overall computational cost.}

\begin{figure*}[htbp]    
    \centering
    \includegraphics[width=\textwidth]{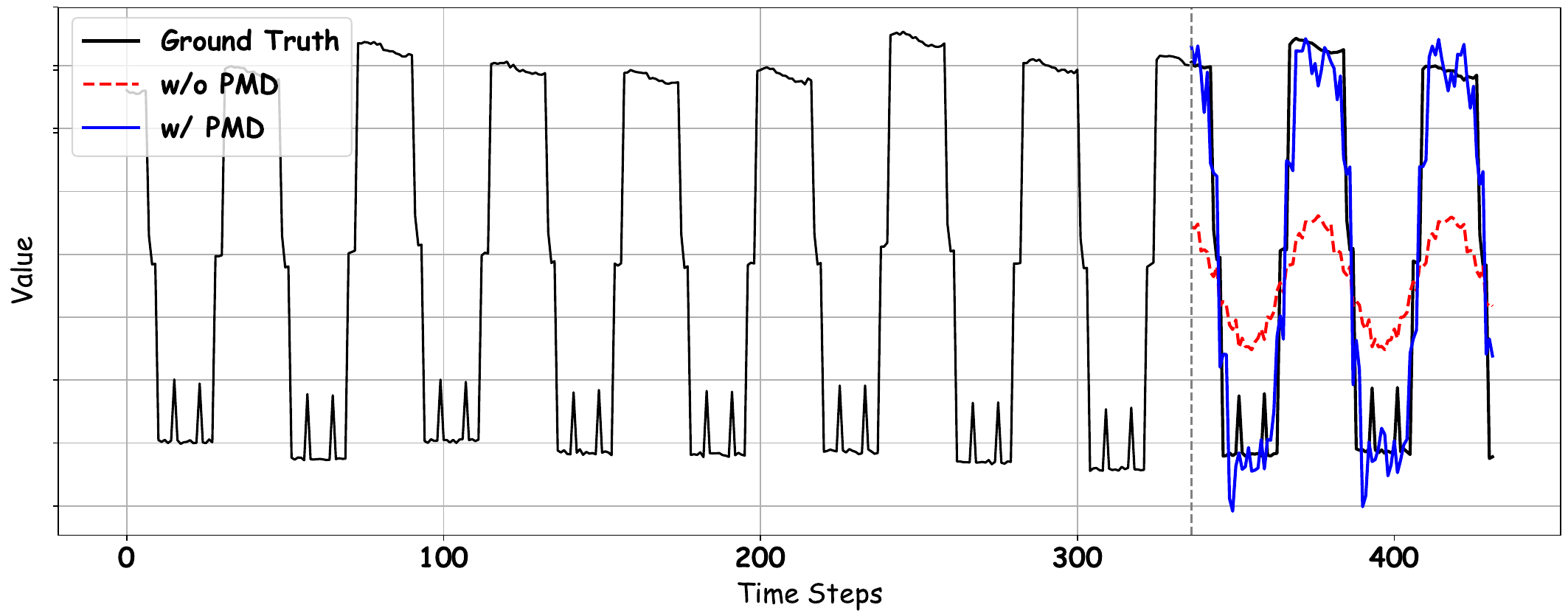}
\caption{{Comparison on synthetic data. The ground truth alternates between pulse and sine shapes with varying scales. The 'w/o PMD' yields smoothed outputs and struggles to recognize the shape similarity, while 'w/ PMD' effectively fits the shapes and trends.}}\label{fig:synthetic_data}
\end{figure*}

\subsection{Comparison on Synthetic Data}
{To further validate the effectiveness of our PMD module, we conduct an experiment on a synthetic dataset. This dataset consists of patches alternating between two different shapes: a sharp pulse wave with large amplitude and a smooth sine wave with small amplitude. To simulate non-stationary time series, the patches exhibit varying scales and are augmented with moderate noise. We compare a standard patch-based Transformer (w/o PMD) against our model incorporating the patch-mean decoupling module (w/ PMD). As illustrated in Figure~\ref{fig:synthetic_data}, the 'w/o PMD' model struggles to recognize true shape similarities due to scale differences between patches, leading to predictions that resemble mostly smooth curves with inadequate trend fitting. In contrast, our 'w/ PMD' model, by removing scale factors, enables attention to focus more effectively on intrinsic shapes, resulting in predictions that better capture both the underlying patterns and long-range trends.}


\section{ON THE USE OF LARGE LANGUAGE MODELS}
{The authors used large language models (LLMs) exclusively for language polishing and minor rephrasing during the final writing stage. All scientific content, ideas, and initial drafts were created entirely by the authors without any text improved by LLMs was carefully checked and edited by the authors. LLMs played no role in developing research questions, designing experiments, analyzing results, or any other aspect of the research itself.}

\end{document}